\documentclass{svjour3}
\pdfoutput=1

\usepackage{courier}
\usepackage[T1]{fontenc}
\usepackage{mathptmx}
\usepackage{textcomp}
\usepackage{times}
\usepackage[scaled=0.92]{helvet}
\usepackage[scaled]{couriers}
\usepackage{multicol}

\usepackage[ruled, linesnumbered]{algorithm2e}
\usepackage{alltt,theorem, amssymb,eucal,latexsym, array, float, graphicx, color}
\usepackage{pgf}
\usepackage[noend]{algorithmic}
\usepackage{epsfig, amsmath}
\usepackage{longtable}
\usepackage{afterpage}
\usepackage{fancyvrb}
\usepackage{listings}

\usepackage{float}
\floatstyle{boxed}
\restylefloat{figure}

\usepackage[colorlinks=true,linkcolor=black]{hyperref}

\newcommand{\ignore}[1]{}

  {\end{alltt}}

\newlength{\halftextwidth}
\newlength{\thirdtextwidth}
\newcommand{\mtrue}{\mathit{true}}

\newtheorem{nn}{Notes}

\newcommand{\myqed}{\mbox{\rule{1.8mm}{1.8mm}}}

\restylefloat{table}

\newcounter{myBackup}

\DefineVerbatimEnvironment{Zinc}{Verbatim}
    {samepage,frame=single,framesep=0.25em,xleftmargin=0.15em,xrightmargin=0.15em}
\DefineVerbatimEnvironment{Shell}{Verbatim}
    {xleftmargin=0.15em,commandchars=\\\{\}}

\setlongtables

\def\CPP{\leavevmode\textrm{\hbox{C\hskip
-0.1ex\raise 0.5ex\hbox{\tiny ++}}}}

\newcommand{\pparagraph}[1]{\paragraph{\textnormal{\textbf{#1.}}}}

\DefineVerbatimEnvironment{Zinc}{Verbatim}
    {samepage,frame=single,framesep=0.25em,xleftmargin=0.15em,xrightmargin=0.15em}

\newcommand{\fig}[3]
  {\begin{figure*}[btp]#3
      \caption{#2} \label{#1} \end{figure*}}

\newcommand{\ands}[1]{\textsf{and}(#1)}
\newcommand{\ors}[1]{\textsf{or}(#1)}
\newcommand{\portfolio}[1]{\textsf{portfolio}(#1)}

\newcommand{\lets}[3]{\textsf{let}(#1,#2,#3)}
\newcommand{\assign}[2]{\textsf{assign}(#1,#2)}
\newcommand{\post}[1]{\textsf{post}(#1)}
\newcommand{\postc}[2]{\textsf{post}(#1,#2)}
\newcommand{\restartf}{\textsf{restart}}
\newcommand{\restart}[2]{\restartf(#1,#2)}

\newcommand{\failure}{\textsf{prune}}

\newcommand{\search}[1]{\textsf{#1}}

\begin{document}
\lstset{basicstyle=\ttfamily,columns=fixed}

\journalname{}
\date{}

\title{Search Combinators}
 \author{Tom Schrijvers
 \and Guido Tack
 \and Pieter Wuille
 \and Horst Samulowitz
 \and Peter J. Stuckey
 }

\institute{
 Tom Schrijvers \and Pieter Wuille \at
  Universiteit Gent, Belgium\\\email{\{tom.schrijvers,pieter.wuille\}@ugent.be}
 \and
 Guido Tack \at
  Monash University, Victoria, Australia\\\email{guido.tack@monash.edu}
 \and
 Pieter Wuille \at
  Katholieke Universiteit Leuven, Belgium\\\email{pieter.wuille@cs.kuleuven.be}
 \and
 Horst Samulowitz \at
  IBM Research, New York, USA\\\email{samulowitz@us.ibm.com}
 \and
 Peter J. Stuckey \at
  National ICT Australia (NICTA) and
 University of Melbourne, Victoria, Australia\\\email{pjs@cs.mu.oz.au}
}

\def\makeheadbox{}
\maketitle

\begin{abstract}
  The ability to model search in a constraint solver can be an essential asset
  for solving combinatorial problems. However, existing infrastructure for
  defining search heuristics is often inadequate. Either modeling capabilities
  are extremely limited or users are faced with a general-purpose programming
  language whose features are not tailored towards writing search heuristics.
  As a result, major improvements in performance may remain unexplored.

  This article introduces \textit{search combinators}, a lightweight and solver-independent 
  method that bridges the gap between a conceptually simple modeling language for search
  (high-level, functional and naturally compositional) and
  an efficient implementation (low-level, imperative and highly non-modular).
  By allowing the user to define application-tailored search strategies from a small
  set of primitives, search combinators effectively provide a rich
  \textit{domain-specific language} (DSL) for modeling search to the user.
  Remarkably, this DSL comes at a low implementation cost to the developer of a constraint solver. 

  The article discusses two modular implementation approaches and shows, by empirical evaluation, that search combinators can be implemented without overhead compared to a native, direct implementation in a constraint solver.
\end{abstract}

\section{Introduction}
\label{sec:intro}
Search heuristics often make all the difference between effectively solving
a combinatorial problem and utter failure. Heuristics make a search algorithm
efficient for a variety of reasons, e.g., incorporation of domain
knowledge, or randomization to avoid heavy-tailed runtimes. Hence, the
ability to swiftly design search heuristics that are tailored towards a problem
domain is essential for performance. This article introduces search
combinators, a versatile, modular, and efficiently implementable language for expressing search heuristics.

\subsection{Status Quo}

In CP, much attention has been devoted to facilitating the modeling of
combinatorial problems. A range of high-level modeling languages, such as
Zinc~\cite{zinc}, OPL~\cite{oplsearch} or Comet~\cite{comet}, enable quick
development and exploration of problem models. However, we see very little
support on the side of formulating accompanying search heuristics. Most
languages and systems, e.g.\ MiniZinc~\cite{minizinc}, Comet~\cite{comet},
Gecode~\cite{gecode}, or ECLiPSe~\cite{eclipse} provide a small set of
predefined heuristics ``off the shelf''. Some systems also support user-defined search based
on a general-purpose programming language (e.g., all of the above
systems except \mbox{MiniZinc}).
The former is clearly too confining, while the latter leaves to be desired in terms of productivity, since implementing a search heuristic quickly becomes a non-negligible effort. This also explains why the set of predefined heuristics is typically small: it takes a lot of time for CP system developers to implement heuristics, too -- time they would much rather spend otherwise improving their system.

\subsection{Contributions}

In this article we show how to resolve this stand-off between solver
developers and users, by introducing a domain-specific modular search language based on combinators, as well as a modular, extensible implementation architecture.

\begin{description}
\item[\textbf{For the user,}] we provide a modeling language for expressing complex search heuristics based on an (extensible) set of primitive
combinators. Even if the users are only provided with a small set of
combinators, they can already express a vast range of combinations. Moreover, using combinators to program application-tailored search is vastly more
productive than resorting to a general-purpose language. 

\item[\textbf{For the system developer,}] we show how to design and implement modular combinators. The modularity of the language thus carries over directly to modularity of the implementation.
Developers do not have to cater explicitly for all possible combinator
combinations. Small implementation efforts result in providing the user with a lot of expressive power.  Moreover, the cost of adding one more
combinator is small, yet the return in terms of additional expressiveness can be quite large.
\end{description}

The technical challenge is to bridge the gap between a conceptually simple search language and an
efficient implementation, which is typically low-level, imperative and highly
non-modular. This is where existing approaches fail; they restrict the
expressiveness of their search specification language to face up to
implementation limitations, or they raise errors when the user strays out of
the implemented subset.

The contribution is therefore the novel design of an expressive, high-level, compositional search language with an equally modular, extensible, and efficient implementation architecture.

\subsection{Approach}

We overcome the modularity challenge by implementing the primitives of our
search language as \emph{mixin} components~\cite{cook}. As in Aspect-Oriented Programming~\cite{kiczales97aspectoriented},
mixin components neatly
encapsulate the \textit{cross-cutting behavior} of primitive search concepts, which are highly entangled in conventional approaches. Cross-cutting means that a mixin component can interfere with the behavior of its sub-components (in this case, sub-searches). The 
combination of encapsulation \textit{and} cross-cutting behavior is essential
for systematic reuse of search combinators.
Without this degree of modularity, minor modifications 
require rewriting from scratch.

An added advantage of mixin components
is extensibility. We can add new features to the language
by adding more mixin components. The cost of adding such a new component is
small, because it does not require changes to the existing ones. Moreover,
experimental evaluation bears out that this modular approach has no significant
overhead compared to the traditional monolithic approach.  Finally, our
approach is solver-independent and therefore makes search combinators a
potential standard for designing search.  

\subsection{Plan of the Article}

The rest of the article is structured as follows. The next section defines the high-level search language in terms of basic heuristics and combinators. Sect.~\ref{sec:modular_combinator_design} shows how the modular language is mapped to a \emph{modular design} of the combinator implementations. Sect.~\ref{sec:modular_combinator_implementation} presents two concrete implementation approaches for combinators and gives an overview of how we integrate search combinators into the MiniZinc toolchain. Sect.~\ref{sec:experiments} verifies that combinators can be implemented with low overhead. Finally, Sect.~\ref{sec:related_work} discusses related approaches, and Sect.~\ref{sec:conclusion} concludes the article.

\subsection{Note to Reviewers}

This article is an extended version of a paper~\cite{schrijvers:2011:0:search} that appeared in the proceedings of the 17th International Conference on Principles and Practice of Constraint Programming (CP) 2011. That paper further developed the ideas laid out in our earlier paper~\cite{samulowitz:2010:0:towards}, which was presented at ModRef 2010.

Compared to the CP'11 conference version, this article features a completely rewritten and more detailed introduction of the combinator language, both from the high-level (Sect.~\ref{sec:highlevel}) and the implementation-level (Sect.~\ref{sec:modular_combinator_design}) point of view. We have completed the integration into MiniZinc and present a full toolchain in Sect.~\ref{sec:minizinccomb}. In addition, the discussion of related work (Sect.~\ref{sec:related_work}) has been extended with much more detail.

\section{High-Level Search Language}
\label{sec:highlevel}

\fig{f:syntax}{Catalog of primitive search heuristics and combinators}{

\begin{tabular}{rcl@{\hspace{3mm}}cl}
  $s$ & $::=$  &  $\failure{}$  & $\mid$ &  $\textsf{ifthenelse}(c,s_1,s_2)$ \\
  & & \mbox{prunes the node} & & \mbox{perform $s_1$ until $c$ is false, then perform $s_2$}\\
  & $\mid$ &  $\textsf{base\_search}(\mathit{vars},\mathit{var}\text{-}\mathit{select},\mathit{domain}\text{-}\mathit{split})$ & $\mid$ &  $\ands{[s_1,s_2,\ldots,s_n]}$ \\
  & & \mbox{label} & & \mbox{perform $s1$, on success $s2$ otherwise fail, \ldots}\\
  & $\mid$ &  $\lets{v}{e}{s}$ & $\mid$ &  $\ors{[s_1,s_2,\ldots,s_n]}$\\
  & & \mbox{introduce new variable $v$ with} & & \mbox{perform $s1$, on termination start $s2$, \ldots} \\
  & & \mbox{initial value $e$, then perform $s$} & $\mid$ &  $\portfolio{[s_1,s_2,\ldots,s_n]}$\\
  & $\mid$ &  $\assign{v}{e}$ & & \mbox{perform $s1$, if not exhaustive start $s2$, \ldots} \\
  & & \mbox{assign $e$ to variable $v$ and succeed} & $\mid$ &  $\restartf(c, s)$ \\
  & $\mid$ &  $\postc{c}{s}$ & & \mbox{restart $s$ as long as $c$ holds}\\
  & & \mbox{post constraint $c$ at every node during $s$} & & 
\end{tabular}
}

This section introduces the syntax of our high-level search language and illustrates its expressive power and modularity by means of examples. The rest of the article then presents an architecture that maps the modularity of the language down to the implementation level.

The search language is used to define a \emph{search heuristic}, which a \emph{search engine} applies to each node of the search tree. For each node, the heuristic determines whether to continue search by creating child nodes, or to prune the tree at that node. The queuing strategy, i.e., the strategy by which new nodes are selected for further search (such as depth-first traversal), is determined separately by the search engine, it is thus orthogonal to the search language. The search language features a number of primitives, listed in the catalog of
Fig.~\ref{f:syntax}. These are the building blocks in terms of which more complex heuristics can be defined, and they can be grouped into \emph{basic heuristics} (\textsf{base\_search} and \textsf{prune}), \emph{combinators} (\textsf{ifthenelse}, \textsf{and}, \textsf{or}, \textsf{portfolio}, and \textsf{restart}), and \emph{state management} (\textsf{let}, \textsf{assign}, \textsf{post}). This section introduces the three groups of primitives in turn.

We emphasize that this catalog is open-ended; we will see that the language
implementation explicitly supports adding new primitives. 

The concrete syntax we chose for presentation uses simple nested terms, which makes it compatible with the \emph{annotation} language of
MiniZinc~\cite{minizinc}. Sect.~\ref{sec:minizinccomb} discusses our implementation of MiniZinc with combinator support.
However, other concrete syntax
forms are easily supported (e.g., we support  \CPP{} and Haskell).

\subsection{Basic Heuristics}

Let us first discuss the two basic primitives, \textsf{base\_search} and \textsf{prune}.

\pparagraph{\textsf{base\_search}}

The most widely used method for specifying a basic heuristic for a constraint
problem is to define it in terms of a \emph{variable selection} strategy which
picks the next variable to constrain, and a \emph{domain splitting} strategy
which splits the set of possible values of the selected variable into two (or
more) disjoint sets. Common variable selection strategies are:
\begin{itemize}
\item \search{firstfail}: select the variable with the smallest current domain,
\item \search{smallest}: select the variable which can take the smallest possible
  value,
\item \search{domwdeg}~\cite{domwdeg}: 
select the variable with smallest ratio of size of current domain and number
  of failures the variable has been involved in, and
\item \search{impact}~\cite{impact}: 
select the variable that will (based on past experience)
reduce the raw search
space of the problem the most.
\end{itemize}
Common domain splitting strategies are:
\begin{itemize}
\item \search{min}: set the variable to its minimum value or greater than its
  minimum,
\item \search{max}: set the variable to its maximum value or less than its
  maximum,
\item \search{median}: set the variable to its median value, or not equal to this
  value, and
\item \search{split}: constrain the variable to the lower half of its range of possible
  values, or its upper half.
\end{itemize}

The CP community has spent a considerable amount of work on defining and
exploring the above and many other variable selection and domain splitting
heuristics. The provision of a flexible language for defining new basic
searches is an interesting problem in its own right, but in this article we
concentrate on search combinators that combine and modify basic searches.

To this end, our search language provides the primitive
\textsf{base\_search}$($$\mathit{vars}$, \emph{var-select},
\emph{domain-split}$)$, which specifies a systematic search. If any of the
variables $\mathit{vars}$ are still not fixed at the current node, it creates
child nodes according to \emph{var-select} and \emph{domain-split} as variable
selection and domain splitting strategies respectively.

Note that \textsf{base\_search} is a CP-specific primitive; other kinds of
solvers provide their own search primitives.  The rest of the search language
is essentially solver-in\-de\-pen\-dent. While the solver provides few basic
heuristics, the search language adds great expressive power by allowing these
to be combined arbitrarily using combinators.

\pparagraph{\failure{}}
The second basic primitive, \failure{}, simply cuts the search tree below the current node. Obviously, this primitive is useless on its own, but we will see shortly how \failure{} can be used together with combinators.

\subsection{Combinators}
\label{Sec::Combinators}
The expressive power of the search language relies on combinators, which combine search heuristics (which can be basic or themselves constructed using combinators) into more complex heuristics.

\pparagraph{\textsf{and}/\textsf{or}}

Probably the most widely used combination of heuristics is \emph{sequential
  composition}. For instance, it is often useful to first label one set of
problem variables before starting to label a second set. The following
heuristic uses the \textsf{and} combinator to first label all the
$\mathit{xs}$ variables using a first-fail strategy, followed by the
$\mathit{ys}$ variables with a different strategy:

\vspace{2pt}
\noindent
\framebox[\columnwidth]{
\begin{minipage}{\columnwidth}
\begin{tabbing}
$\textsf{and}([$\=$\textsf{base\_search}(\mathit{xs},\mathrm{firstfail},\mathrm{min}),$\\
\>$\textsf{base\_search}(\mathit{ys},\mathrm{smallest},\mathrm{max})])$
\end{tabbing}
\end{minipage}
}
\vspace{2pt}

As you can see in Fig.~\ref{f:syntax}, the \textsf{and} combinator accepts a list of searches $s_1,\dots,s_n$, and performs their and-sequential composition. And-sequential means, intuitively, that solutions are found by performing \emph{all} the sub-searches sequentially down one branch of the search tree, as illustrated in Fig.~\ref{f:combdiag}.1.

The dual combinator, $\textsf{or}([s_1,\dots,s_n])$, performs a disjunctive combination of its sub-searches -- a solution is found using \emph{any} of the sub-searches (Fig.~\ref{f:combdiag}.2), trying them in the given order.

\fig{f:combdiag}{Primitive combinators}{
\vspace{2pt}
\hspace{.2\textwidth}\includegraphics[width=.6\textwidth]{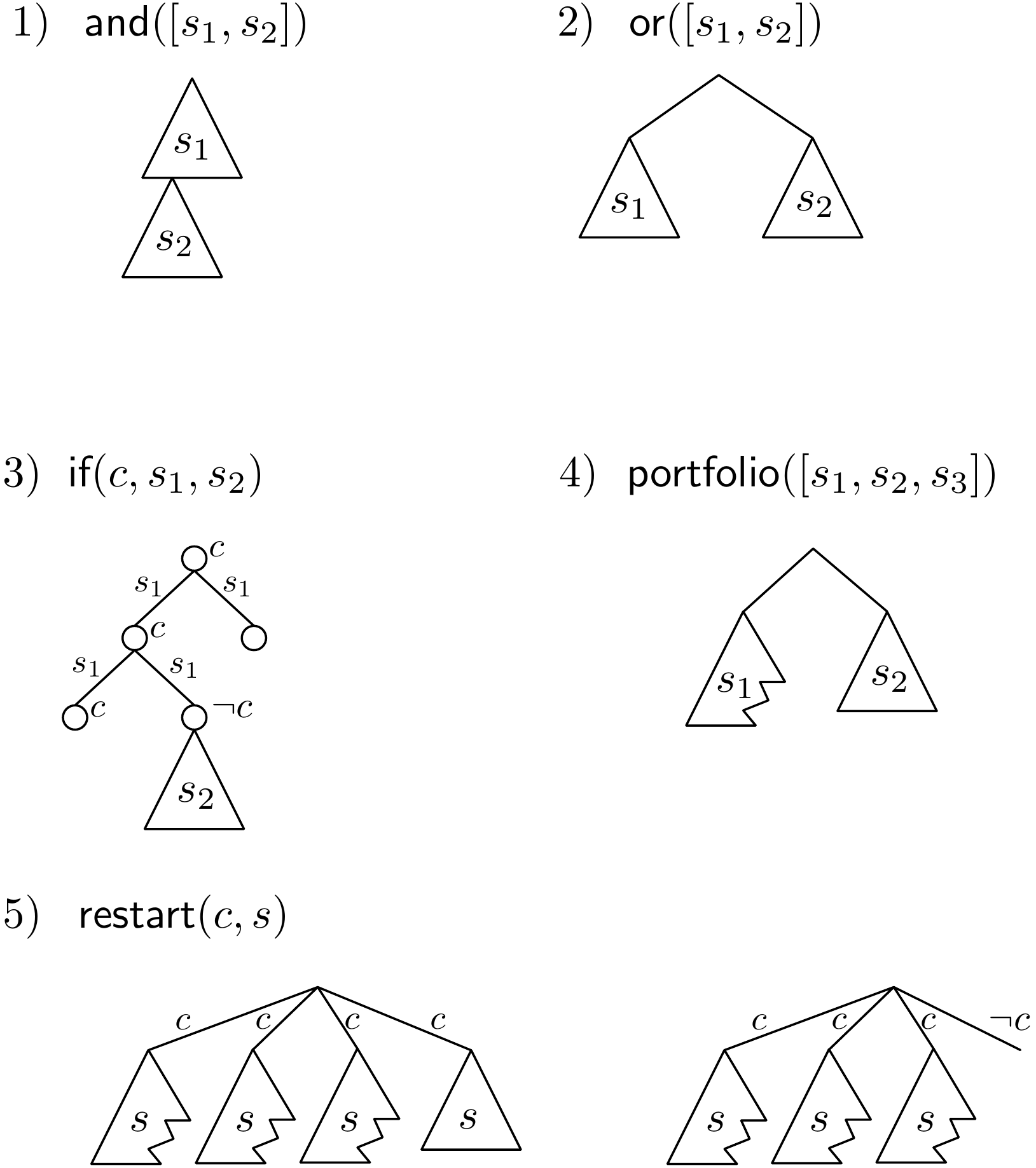}  
\vspace{2pt}
}

\pparagraph{Statistics and \textsf{ifthenelse}}

The \textsf{ifthenelse} combinator is centered around a conditional expression $c$. As long as $c$ is true for the current node, the sub-search $s_1$ is used. Once $c$ is false, $s_2$ is used for the complete subtree below the current node (see Fig.~\ref{f:combdiag}.3).

We do not specify the \textit{expression language} for conditions in detail, we simply assume that it comprises the typical arithmetic and comparison
operators and literals that require no further explanation. It is notable though that the language can refer to the constraint variables and parameters of
the underlying model. Additionally, a condition may refer to one or
more \textit{statistics} variables.
Such statistics are collected for the duration of a subsearch until the
condition is met. For instance $\textsf{ifthenelse}(\textsf{depth} < 10,s_1,s_2)$
maintains the search \textsf{depth} statistic during subsearch $s_1$. At depth 10,
the \textsf{ifthenelse} combinator switches to subsearch $s_2$.

We distinguish two forms of statistics: 
\textit{Local statistics} such as
\textsf{depth} and \textsf{discrepancies} express properties of individual nodes. \textit{Global statistics} such as number of explored \textsf{nodes}, encountered \textsf{failures},  \textsf{solution}, and \textsf{time}
are computed for entire search trees.

It is worthwhile to mention that developers (and advanced users)
can also define their own statistics,
just like combinators, to complement any predefined ones. In fact, Sect.~\ref{sec:modular_combinator_design} will show that statistics can be implemented as a \textit{subtype} of combinators that can be queried for the statistic's value. 

\pparagraph{Abstraction}

Our search language draws its expressive power from the combination of
primitive heuristics using combinators. An important aspect of the search language is \emph{abstraction}: the ability to create new combinators by effectively defining macros in terms of existing combinators.

For example, we can define the limiting combinator $\textsf{limit}(c,s)$ to perform $s$ while condition $c$ is satisfied, and otherwise cut the search tree using \failure{}:

\vspace{2pt}
\noindent
\framebox[\columnwidth]{
\begin{minipage}{\columnwidth}
\begin{tabbing}
$\textsf{limit}(c,s) \equiv \textsf{ifthenelse}(c,s,\failure{})$
\end{tabbing}
\end{minipage}
}

\vspace{2pt}
\noindent
The well-known $\textsf{once}(s)$ combinator is a special case of the limiting
combinator where the number of solutions is less than one. This is simply achieved by maintaining and accessing the \textsf{solutions} statistic:

\vspace{2pt}
\noindent
\framebox[\columnwidth]{
\begin{minipage}{\columnwidth}
\begin{tabbing}
$ \textsf{once}(s) \equiv \textsf{limit}(\textsf{solutions} < 1, s)$
\end{tabbing}
\end{minipage}
}

\pparagraph{Exhaustiveness and \textsf{portfolio}/\textsf{restart}}

The behavior of the final two combinators, \textsf{portfolio} and \textsf{restart}, depends on whether their sub-search was \emph{exhaustive}. Exhaustiveness simply means that the search has explored the entire subtree without ever invoking the \failure{} primitive.

The $\textsf{portfolio}([s_1,\dots,s_n])$ combinator performs $s_1$ until it has explored the whole subtree. If $s_1$ was exhaustive, i.e., if it did not call \failure{} during the exploration of the subtree, the search is finished. Otherwise, it continues with $\textsf{portfolio}([s_2,\dots,s_n])$. This is illustrated in Fig.~\ref{f:combdiag}.4, where the subtree of $s_1$ represents a non-exhaustive search, $s_2$ is exhaustive and therefore $s_3$ is never invoked.

An example for the use of \textsf{portfolio} is the $\textsf{hotstart}(c,s_1,c_2)$ combinator. It performs search heuristic $s_1$
while condition $c$ holds to initialize global
parameters for a second search $s_2$. This heuristic can for example be used to initialize the widely applied \emph{Impact} heuristic \cite{impact}.
Note that we assume here that the parameters to be initialized are maintained by the underlying solver, so we omit an explicit reference to them.

\vspace{2pt}
\noindent
\framebox[\columnwidth]{
\begin{minipage}{\columnwidth}
\begin{tabbing}
xx \= xx \= xx \= xx \= xx \= \kill
$\textsf{hotstart}(c,s_1,s_2) \equiv$ $\portfolio{[\textsf{limit}(c, s_1),s_2]}$
\end{tabbing}
\end{minipage}
}
\vspace{2pt}

The $\textsf{restart}(c,s)$ combinator repeatedly runs $s$ in full. If $s$ was not exhaustive, it is restarted, until condition $c$ no longer holds. Fig.~\ref{f:combdiag}.5 shows the two cases, on the left terminating with an exhaustive search $s$, on the right terminating because $c$ is no longer true.

The following implements random restarts, where search is stopped after 1000 failures and restarted with a random strategy:

\vspace{2pt}
\noindent
\framebox[\columnwidth]{
\begin{minipage}{\columnwidth}
\begin{tabbing}
$\restart{\mathrm{true}}{\textsf{limit}(\textsf{failures}<1000,\textsf{base\_search}(\mathit{xs},\mathrm{randomvar},\mathrm{randomval}))}$
\end{tabbing}
\end{minipage}
}

\vspace{2pt}
\noindent
Clearly, this strategy has a flaw: If it takes more than 1000 failures to find the solution, the search will never finish. We will shortly see how to fix this by introducing user-defined search variables.

The \failure{} primitive is the only source of non-exhaustiveness.
Combinators propagate exhaustiveness in the obvious way:
\begin{itemize}
  \item $\ands{[s_1,\ldots,s_n]}$ is exhaustive if all $s_i$ are
  \item $\ors{[s_1,\ldots,s_n]}$ is exhaustive if all $s_i$ are
  \item $\portfolio{[s_1,\ldots,s_n]}$ is exhaustive if one $s_i$ is
  \item $\restart{c}{s}$ is exhaustive if the last iteration is exhaustive
  \item $\textsf{ifthenelse}(c,s_1,s_2)$ is exhaustive if $s_1$ and $s_2$ are
\end{itemize}

\subsection{State Access and Manipulation}
\label{sec:stateaccess}

The remaining three primitives, \textsf{let}, \textsf{assign}, and \textsf{post}, are used to access and manipulate the state of the search:
\begin{itemize}
\item $\lets{v}{e}{s}$ introduces a new search variable $v$
  with initial value of the expression $e$ and visible in the search $s$, then 
continues with $s$. Note that search variables are distinct from the decision variables of the model.
\item $\assign{v}{e}$: assigns the value of the expression $e$ to search variable $v$
  and succeeds.
\item
$\postc{c}{s}$: provides access to the underlying constraint solver, posting a constraint $c$
at every node during $s$. If $s$ is omitted, it posts the constraint and immediately succeeds.
\end{itemize}

These primitives add a great deal of expressivity to the language, as the following examples demonstrate.

\paragraph{Random restarts:}
Let us reconsider the example using random restarts from the previous section, which suffered from incompleteness because it only ever explored 1000 failures. A standard way to make this strategy complete is to increase the limit geometrically with each iteration:

\vspace{2pt}
\noindent
\framebox[\columnwidth]{
\begin{minipage}{\columnwidth}
\begin{tabbing}
$\textsf{geom\_restart}(s) \equiv$
$\textsf{let}($\=$\textit{maxfails},100,$\\
\>$\restart{\mathrm{true}}{
\portfolio{[$\=$\textsf{limit}(\textsf{failures}<\textit{maxfails},s),$\\
\>\>$\assign{\textit{maxfails}}{\textit{maxfails}*1.5},$\\
\>\>$\failure]}}$
\end{tabbing}
\end{minipage}
}
\vspace{2pt}

\noindent
The search initializes the search variable $\mathit{maxfails}$ to $100$, and then calls search $s$ with $\mathit{maxfails}$ as the limit. If the search is exhaustive, both the \textsf{portfolio} and the \textsf{restart} combinators are finished. If the search is not exhaustive, the limit is multiplied by $1.5$, and the search starts over. Note that \textsf{assign} succeeds, so we need to call \failure{} afterwards in order to propagate the non-exhaustiveness of $s$ to the \restartf{} combinator.

\paragraph{Branch-and-bound:}
A slightly more advanced example is the branch-and-bound optimization strategy:

\vspace{2pt}
\noindent
\framebox[\columnwidth]{
\begin{minipage}{\columnwidth}
\begin{tabbing}
$\textsf{bab}(\mathit{obj},s) \equiv$ $\textsf{let}(\mathit{best},\infty,$
$\textsf{post}(\mathit{obj} < \mathit{best},$\= $\textsf{and}([$\= $s, \assign{\mathit{best}}{\mathit{obj}}])))$
\end{tabbing}
\end{minipage}
}

\vspace{2pt}
\noindent
It introduces a variable $\mathit{best}$ that initially takes value
$\infty$ (for minimization). In every node, it posts a constraint to bound the objective variable by \textit{best}. Whenever a new solution is found, the bound is updated accordingly using \textsf{assign}.

The \textsf{bab} example demonstrates how search variables (like $\mathit{best}$) and model variables (like $\mathit{obj}$) can be mixed in expressions. This makes it possible to remember the state of the search between invocations of a heuristic. All of the following combinators make use of this feature.

\paragraph{Restarting branch-and-bound:}
This is a twist on regular 
branch-and-bound that restarts whenever a solution is found.

\vspace{2pt}
\noindent
\framebox[\columnwidth]{
\begin{minipage}{\columnwidth}
\begin{tabbing}
$\textsf{restart\_bab}(\mathit{obj},s) \equiv$ 
$\textsf{let}(\mathit{best},\infty,$ \restartf($\mtrue$, \=\textsf{and}([\= $\post{\mathit{obj} <
  \mathit{best}}, \textsf{once}(s),$ \\
\> \>  $\assign{\mathit{best}}{\mathit{obj}}])))$
\end{tabbing}
\end{minipage}
}

\paragraph{Radiotherapy treatment planning:}
The following search heuristic can be used to solve radiotherapy treatment
planning problems~\cite{radiation}. The heuristic minimizes a variable $k$
using branch-and-bound (\textsf{bab}), first searching the variables $N$,
and then verifying the solution by partitioning the problem along the
$\mathit{row}_{i}$ variables for each row $i$ one at a time (expressed as a MiniZinc array comprehension).
Failure on one row
must be caused by the search on the variables in $N$, 
and consequently search never backtracks into other rows. 

This behavior is similar to the \textsf{once} combinator defined above. However, when a single solution is found, the search should be considered exhaustive. We therefore need an exhaustive variant of \textsf{once}, which can be implemented by replacing \failure{} with $\textsf{post}(\textit{false})$: \\
\framebox[\columnwidth]{
\begin{minipage}{\columnwidth}
\begin{tabbing}
$ \textsf{exh\_once}(s) \equiv \textsf{ifthenelse}(\textsf{solutions} < 1, s, \post{\textit{false}})$
\end{tabbing}
\end{minipage}
}

\vspace{2pt}
\noindent
This allows us to express the entire search strategy for radiotherapy treatment planning:

\vspace{2pt}
\noindent
\framebox[\columnwidth]{
\begin{minipage}{\columnwidth}
\begin{tabbing}
$\textsf{bab}($\=$\mathit{k},$
    $\textsf{and}($\=$[\textsf{base\_search}(N,\dots)]$\texttt{++}\\
    \>\> $[\textsf{exh\_once}(\textsf{base\_search}($\=$\mathit{row}_i,\dots))$ $|$ $i$ in $1..n]))$ 
\end{tabbing}
\end{minipage}
}

\paragraph{For:}
The \textsf{for} loop construct ($v\in[l,u]$) can be defined as:

\vspace{2pt}
\noindent
\framebox[\columnwidth]{
\begin{minipage}{\columnwidth}
\begin{tabbing}
\textsf{for}$(v,l,u,s)  \equiv$ \= \textsf{let}$(v, l,\restartf($\= $v \leq u,$\\ 
\> \> $\portfolio{[s,\textsf{and}([\assign{v}{v+1},\failure{}])]))}$
\end{tabbing}
\end{minipage}
}

\vspace{2pt}
\noindent
It simply runs $u-l+1$ times the search $s$, which of course is only sensible if $s$
makes use of side effects or the loop variable $v$. As in the \textsf{geom\_restart} combinator above, \failure{} propagates the non-exhaustiveness of $s$ to the \restartf{} combinator.

\paragraph{Limited discrepancy search~\emph{\cite{lds}}}
with an upper limit of $l$ discrepancies
for an underlying search $s$.

\vspace{2pt}
\noindent
\framebox[\columnwidth]{
\begin{minipage}{\columnwidth}
\begin{tabbing}
\textsf{lds}$(l,s) \equiv$ \= 
      \textsf{for}$($\=$n$, $0$, $l$,
      $\textsf{limit}$(\=$\textsf{discrepancies}\leq n,s))$ 
\end{tabbing}
\end{minipage}
}

\vspace{2pt}
\noindent
The \textsf{for} construct iterates the maximum number of discrepancies $n$ from 0 to $l$, while \textsf{limit} executes $s$ as long as the number of discrepancies is smaller than $n$. The search makes use of the
\textsf{discrepancies} statistic that is maintained by the search
infrastructure. The original LDS~\cite{lds} 
visits the nodes in a specific order. The search described here visits the same nodes in the same order of discrepancies, but possibly in a different individual order -- as this is determined by the global queuing strategy.

The following is a combination of branch-and-bound and limited discrepancy search for solving job shop scheduling problems, as described in~\cite{lds}. The heuristic searches the Boolean variables $\mathit{prec}$, which determine the order of all pairs of tasks on the same machine. As the order completely determines the schedule, we then fix the start times using $\textsf{exh\_once}$.

\vspace{2pt}
\noindent
\framebox[\columnwidth]{
\begin{minipage}{\columnwidth}
\begin{tabbing}
$\textsf{bab}($\=$\mathit{makespan},$
   $\textsf{lds}(\infty,\textsf{and}([$\=$\textsf{base\_search}(\mathit{prec},\dots),$\\
   \>\>$\textsf{exh\_once}(\textsf{base\_search}(\mathit{start},\dots))])))$
\end{tabbing}
\end{minipage}
}

\vspace{2pt}
\noindent
Fully expanded, this heuristic consists of 17 combinators and is 11
combinators deep.

\paragraph{Iterative deepening~\emph{\cite{ida}}} for an underlying search $s$ is a
particular instance of the more general pattern of restarting with an
updated bound, which we have already seen in the \textsf{geom\_restart} example. Here, we generalize this idea:

\vspace{2pt}
\noindent
\framebox[\columnwidth]{
\begin{minipage}{\columnwidth}
\begin{tabbing}
$\textsf{id}(s) \equiv \textsf{ir}(\textsf{depth},0,+,1,\infty,s)$ \\ 
$\textsf{ir}(p,l$\=$,\oplus,i,u,s) \equiv$ 
   $\textsf{let}(n,l,$ 
   $\restartf(n \leq u,\textsf{and}([$\=$\assign{n}{n \oplus i},$ \\
                   \>                        \> $\textsf{limit}(p\leq n,s)])))$ 
\end{tabbing}
\end{minipage}
}

\vspace{2pt}
\noindent
With \textsf{let}, bound $n$ is initialized to $l$. Search $s$ is pruned
when statistic $p$ exceeds $n$, but iteratively restarted by \restartf{} with $n$ updated to $n \oplus i$. 
The repetition stops when $n$ exceeds $u$ or when $s$
has been fully explored. The bound increases geometrically, if we supply $*$ for $\oplus$, as in the \textsf{restart\_flip}
heuristic:

\vspace{2pt}
\noindent
\framebox[\columnwidth]{
\begin{minipage}{\columnwidth}
\begin{tabbing}
xx \= xxx \kill
$\textsf{restart\_flip}(p,l,i,u,s_1,s_2) \equiv$
$\textsf{let}(\textit{flip},1,\textsf{ir}(p,l,*,\mathit{i},u,\textsf{and}([$\=$\textsf{assign}(\mathit{flip},1-\mathit{flip}),$
\\
       \> $\textsf{ifthenelse}(\textit{flip}=1, s_1, s_2)])))$
\end{tabbing}
\end{minipage}
}

\vspace{2pt}
\noindent
The \textsf{restart\_flip} search alternates between two search heuristics $s_1$ and $s_2$. Using this as its default strategy in the \emph{free search} category, the lazy clause generation solver \textit{Chuffed} scored most points in the 2010 and 2011 MiniZinc Challenges.\footnote{\url{http://www.g12.csse.unimelb.edu.au/minizinc/challenge2011/}}

\paragraph{Probe search:}
Try out two searches $s_1$ and $s_2$ to a limited extent
defined by condition $c$. Then, for the remainder, use the search
that resulted in the best solution so far. \\
\framebox[\columnwidth]{
\begin{minipage}{\columnwidth}
\begin{tabbing}
xx \= xx \= xx \= xx \= xx \= \kill
$\textsf{probe}(c,\mathit{obj},s_1,s_2) \equiv$ $\textsf{let}(best_1, \infty,$
$\textsf{let}($\= $best_2, \infty,$ \\
\>$\portfolio{[$ \= $\textsf{limit}(c,\textsf{and}([s_1, \textsf{assign}(best_1,\mathit{obj})])}$ \\
     \>  \> $\textsf{limit}(c,\textsf{and}([s_2, \textsf{assign}(best_2,\mathit{obj})]))$ \\
     \>  \> $\textsf{ifthenelse}(best_1 \leq best_2, s_1, s_2)])))$
\end{tabbing}
\end{minipage}
}

\paragraph{Dichotomic search~\emph{\cite{dichotomic}}}
solves an optimization problem by repeatedly partitioning the interval in which the possible optimal solution can lie. It can be implemented by restarting as long the lower bound has not met the upper bound (line 2), computing the middle (line 3), and then using an \textsf{or} combinator to try the lower half (line 5). If it succeeds, $\mathit{obj}-1$ is the new upper bound, otherwise, the lower bound is increased (line 6).

\vspace{2pt}
\noindent
\framebox[\columnwidth]{
\begin{minipage}{\columnwidth}
\begin{tabbing}
  $\textsf{dicho}(s,\mathit{obj},\mathit{lb},\mathit{ub})\equiv$\=
  $\textsf{let}(l,\mathit{lb},\textsf{let}(u,\mathit{ub},$\\
  \>$\textsf{restart}(l<u,$\\
  \>\quad\=$\textsf{let}(h,l+\lceil(u-l)/2\rceil,$\\
  \>\>$\textsf{once}(\textsf{or}([$\\
  \>\>\quad$\textsf{and}([\textsf{post}(l\leq\mathit{obj}\leq h), s, \textsf{assign}(u,\mathit{obj}-1)]),$\\
  \>\>\quad$\textsf{and}([\textsf{assign}(l,h+1),\textsf{prune}])]))$\\
  \>$))))$
\end{tabbing}
\end{minipage}
}

\section{Modular Combinator Design}
\label{sec:modular_combinator_design}

The previous section caters for the user's needs, presenting a high-level
modular syntax for our combinator-based search language.  To cater for the
system developer's needs, this section goes beyond modularity of syntax,
introducing modularity of \emph{design}.

\paragraph{Modularity of design} is the one property that makes our approach practical. Each combinator corresponds to a separate
module that has a meaning and an implementation independent of the other
combinators. This enables us to actually realize the search specifications defined by modular syntax.

Modularity of design also enables growing a system from a small set of
combinators (e.g., those listed in Fig.~\ref{f:syntax}), gradually adding more
as the need arises. Advanced users can complement the
system's generic combinators with a few application-specific ones.

\paragraph{Solver independence} is another notable property of our approach.
While a few combinators access solver-specific functionality (e.g.,
\textsf{base\_search} and \textsf{post}), the approach as such and most
combinators listed in Fig.~\ref{f:syntax} are in fact generic (solver- and
even CP-independent); their design and implementation is reusable.

The solver-independence of our approach is reflected in the minimal interface
that solvers must implement. This interface consists of an abstract type
\texttt{State} which represents a state of the solver (e.g., the variable
domains and accumulated constraint propagators) which supports copying. Truly
no more is needed for the approach or all of the primitive combinators in
Fig.~\ref{f:syntax}, except for \textsf{base\_search} and \textsf{post} which
require CP-aware operations for querying variable domains, solver status
and posting constraints, and possibly interacting with statistics maintained
by the solver.  
Note that there need not be a 1-to-1 correspondence
between an implementation of the abstract \texttt{State} type and the solver's
actual state representation; e.g., for solvers based on trailing, recomputation techniques \cite{Perron99SearchProcedures} can be used. We have implementations of the interface based on both copying and trailing.

\paragraph{}
In the following we explain our design in detail by means of code implementations of most
of the primitive combinators we have covered in the previous section.

\lstdefinelanguage{pseudo} 
  {morekeywords={combinator,local,global,int,if,else,for,each,node,bool,not,and,condition,stack,then,return,abort,success,failure,protocol,post,false,true,top,this,queue,new,while,print}
  ,classoffset=1
  ,morekeywords={enter,start,init,exit,eval}
  ,keywordstyle=\underline
  ,classoffset=0
  ,sensitive=false
  ,morecomment=[l]{//}
  ,morecomment=[s]{/*}{*/}
  ,morestring=[b]"
  ,literate={printc}{{\sffamily print}}4
            {dfs}{{\sffamily dfs}}2
            {portfolio}{{\sffamily portfolio}}5
            {restart}{{\sffamily restart}}5
            {prune}{{\sffamily prune}}5
            {ifthenelse}{{\sffamily ifthenelse}}{7}
            {postc}{{\sffamily post}}4
            {andc}{{\sffamily and}}3
            {limitsolutions}{{\sffamily limitsolutions}}{9}
            {solutionslimit}{{\sffamily solutionslimit}}{9}
            {base_search}{{\sffamily base\_search}}9
  ,
  }

\subsection{The Message Protocol}\label{s:design:protocol}

\begin{figure}[t]
\begin{center}
\pgfimage[width=0.55\columnwidth]{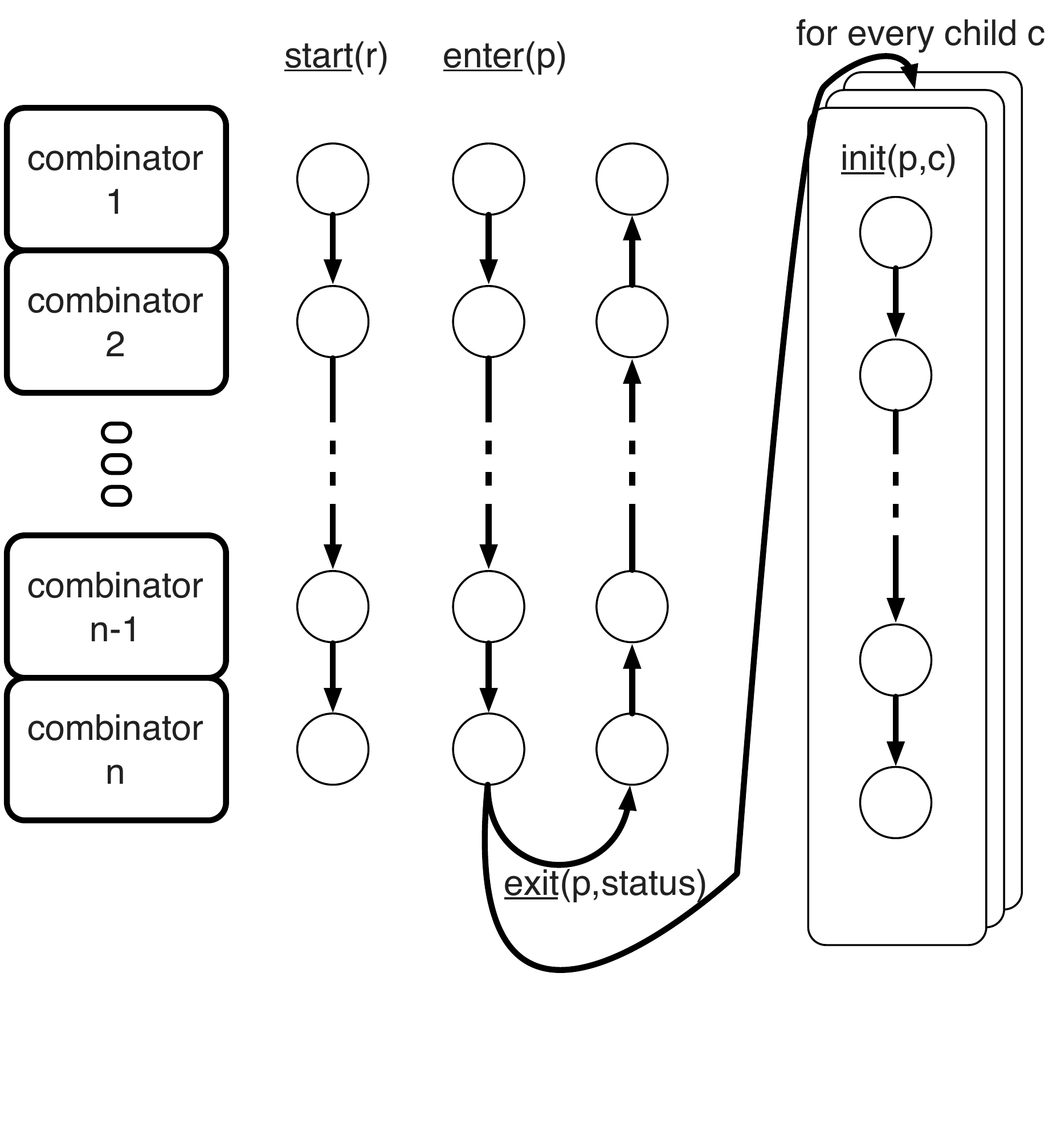}
\end{center}
\vspace{-1cm}
\caption{The modular message protocol}\label{fig:protocol}
\end{figure}

To obtain a modular design of search combinators we step away from the idea
that the behavior of a search combinator, like the \textsf{and} combinator,
forms an indivisible whole; this leaves no room for interaction. The key
insight here is that we must identify finer-grained steps, defining how
different combinators interact at each node in the search tree. Interleaving
these finer-grained steps of different combinators in an appropriate manner
yields the composite behavior of the overall search heuristic, where each
combinator is able to cross-cut the others' behavior.

Considering the diversity of combinators and the fact that not all units of
behavior are explicitly present in all of them, designing this protocol of
interaction is non-trivial. It requires studying the intended behavior and
interaction of combinators to isolate the fine-grained units of behavior and
the manner of interaction. The contribution of this section is an elegant and
conceptually uniform design that is powerful enough to express all the
combinators presented in this article.

We present this design in the form of a \textit{message protocol}. The protocol
specifies a set of messages (i.e., an interface with one procedure for each
fine-grained step) that have to be implemented by all combinators. In pseudo-code,
this protocol for combinators consists of four different messages:

\begin{minipage}{\textwidth}
\begin{lstlisting}[language=pseudo,xleftmargin=3cm]
protocol combinator
  start(rootNode);
  enter(currentNode);
  exit(currentNode,status);
  init(parentNode,childNode);
\end{lstlisting}
\end{minipage}

All of the message signatures specify one or two search tree \textit{nodes} as
parameters. Each such node keeps track of a solver \texttt{State}
and the information associated by combinators to that \texttt{State}. We observe
three different access patterns of nodes:
\begin{enumerate}
\item 
In keeping with the solver independence stipulated above, we will see that most
combinators only query and update their associated information and do not
access the underlying solver \texttt{State} at all. 
\item
Restarting-based combinators, like \lstinline{restart} and \lstinline{portfolio},
copy nodes. This means copying the underlying solver \texttt{State} 
and all associated information. 
\item
Finally, selected solver-specific combinators 
like \lstinline{base_search} do perform solver-specific operations
on the underlying \texttt{State}, like querying variable domains and posting constraints.
\end{enumerate}

In addition to the message signatures, the protocol also stipulates in what
order the messages are sent among the combinators (see
Fig.~\ref{fig:protocol}).  While in general a combinator composition is
tree-shaped, the processing of any single search tree node $p$ only involves a
stack of combinators. For example, given
$\ors{[\ands{[s_1,s_2]},\ands{[s_3,s_4]}]}$, either $s_1,s_2$ or $s_3,s_4$ are
\emph{active} for $p$. The picture shows this stack of active combinators on
the left. Every combinator in the stack has both a \textit{super}-combinator
above and a \textit{sub}-combinator below, except for the \textit{top} and the
\textit{bottom} combinators. The bottom is always a basic heuristic (\textsf{base\_search}, \failure, \textsf{assign}, or \textsf{post}). The important aspect to take away from the picture is the direction
of the four different messages, either top-down or bottom-up.

\subsection{Basic Setup}

\lstset{language=pseudo}

Before we delve into the interesting search combinators, we first present an
example implementation of the basic setup consisting of a base search
(\textsf{base\_search}) and a search engine (\textsf{dfs}). This allows us to
express overall search specifications of the form:\\
\texttt{\textsf{dfs}(\textsf{base\_search}(\textit{vars},\textit{var-select},\textit{domain-split}))}.

\pparagraph{Base Search}
We do not provide full details on a \lstinline{base_search} combinator, as it is
not the focus of this article. However, we will point out the aspects relevant to
our protocol.

In the \lstinline{enter} message, the node's solver state is propagated.
Subsequently, the condition \lstinline{isLeaf(c,vars)} checks whether the
solver state is unsatisfiable or there are no more variables to assign. If
either is the case, the exit status (respectively \lstinline{failure} or
\lstinline{success}) is sent to the \lstinline{parent} combinator. For now,
the \lstinline{parent} combinator is just the search engine, but later we will
see how how other combinators can be inserted between the search engine and the
base search.

If neither is the case, the search branches depending on the
variable selection and domain splitting strategies. This involves creating a child node for each branch,
determining the variable and value for that child and posting the assignment to
the child's state. Then, the \lstinline{top} combinator (i.e., the engine) is
asked to initialize the child node.  Finally the child node is pushed onto the
search queue.
 
\begin{lstlisting}[language=pseudo,frame=single]
combinator base_search(vars,var-select,domain-select)
  enter(c):
    c.propagate
    if isLeaf(c,vars)
      parent.exit(leafstatus(c))
    pos = ...       // from vars based on var-select
    for each child: // based on domain-select
      val = ...     // from values of var based on domain-select
      child.post(vars[pos]=val)
      top.init(c,child)
      queue.push(child)
\end{lstlisting}
Note that, as the \lstinline{base_search} combinator is a base combinator, its
\lstinline{exit} message is immaterial (there is no child heuristic of \lstinline{base_search} that could ever call it). The \lstinline{start} and \lstinline{init}
messages are empty.
Many variants on and generalizations of the above implementation are possible.

\pparagraph{Depth-first search engine} 
The engine \textsf{dfs} serves as a pseudo-combinator at the \lstinline{top} of
a combinator expression \lstinline{heuristic} and serves as the
\lstinline{heuristic}'s immediate parent as well. It maintains the
\lstinline{queue} of nodes, a stack in this case. The search \lstinline{start}s
from a given \lstinline{root} node by starting the \lstinline{heuristic} with that
node and then \lstinline{enter}ing it. Each time a node has been processed, new nodes may have
been pushed onto the queue. These are popped and \lstinline{enter}ed successively.

\begin{lstlisting}[language=pseudo,frame=single]
combinator dfs(heuristic)
  start(root):
    top=this
    heuristic.parent=this
    queue=new stack()
    heuristic.start(root)
    heuristic.enter(root)
    while not queue.empty
      heuristic.enter(queue.pop())

  init(n,c):
    heuristic.init(n,c)
\end{lstlisting}
The engine's \lstinline{exit} message is empty, the \lstinline{enter} message
is never called and the \lstinline{init} message delegates initialization to the
\lstinline{heuristic}.

Other engines may be formulated with different queuing strategies.
\subsection{Combinator Composition} 

The idea of search combinators is to augment a \textsf{base\_search}. We
illustrate this with a very simple \textsf{print} combinator that prints
out every solution as it is found. For simplicity we assume a solution
is just a set of constraint variables \textit{vars} that is supplied as 
a parameter. Hence, we obtain the basic search setup with solution printing
with: 
\[ \texttt{\textsf{dfs}(\textsf{print}(\textit{vars},\textsf{base\_search}(\textit{vars},\textit{strategy})))} \]

\pparagraph{Print}
The \textsf{print} combinator is parametrized by a set of variables
\lstinline{vars} and a search combinator \lstinline{child}. Implicitly, in a
composition, that \lstinline{child}'s \lstinline{parent} is set to the
\textsf{print} instance. The same holds for all following search combinators
with one or more children.

The only message of interest for \textsf{print} is \lstinline{exit}. When the exit
status is \lstinline{success}, the combinator prints the variables and propagates
the message to its parent.
\begin{lstlisting}[language=pseudo,frame=single]
combinator printc(vars,child)
  exit(c,status):
    if status==success 
      print c.vars
    parent.exit(c,status)
\end{lstlisting}
The other messages are omitted. Their behavior is default: they all propagate
to the child. The same holds for the omitted messages of following unary combinators.

\subsection{Binary Combinators}

Binary combinators are one step up from unary ones. They combine two
complete search heuristics into a composite one. 
The most basic binary
combinator is the binary version of 
\lstinline{andc}. For instance, if we need to label
two sets of variables, we can do so with
\[ \texttt{\textsf{and}(\textsf{base\_search}($\mathit{vars}_1$,\dots),\textsf{base\_search}($\mathit{vars}_2$,\dots))} \]
The principle shown here easily generalizes to $n$-ary combinators.

\pparagraph{And} 
The (binary) \lstinline{andc} combinator has two children, \lstinline{left} and
\lstinline{right}. In order to keep track of what child combinator is handling
a particular node, the \lstinline{andc} combinator associates with every node an
\lstinline{inLeft} Boolean variable. The \lstinline{local} keyword indicates
that every node has its own instance of that variable. We denote the instance
of the \lstinline{inLeft} variable associated with node \lstinline{c} as \lstinline{c.inLeft}.

When \lstinline{enter}ing a node, it is delegated to the \lstinline{left} or
\lstinline{right} combinator based on \lstinline{inLeft}.  At the
\lstinline{start}, the \lstinline{root} node is delegated to the
\lstinline{left} combinator, so its \lstinline{inLeft} variable is set to
\lstinline{true}.  The value of \lstinline{inLeft} is inherited in
\lstinline{init} from the current node to its children. Upon a successful
\lstinline{exit} for \lstinline{left}, the leaf node becomes the root of a new
subtree that is further handled by the \texttt{right} combinator.

\begin{lstlisting}[language=pseudo,frame=single]
combinator andc(left,right) {
  local bool inLeft

  start(root):
    root.inLeft=true
    left.start(root)

  enter(c):
    if c.inLeft 
      left.enter(c)
    else 
      right.enter(c)
  
  exit(c,status):
    if c.inLeft and status==success 
      c.inLeft=false
      right.start(c)
      right.enter(c)
    else
      parent.exit(c,status)
  
  init(p,c):
    c.inLeft=p.inLeft
    if c.inLeft 
      left.init(p,c) 
    else 
      right.init(p,c)
\end{lstlisting}

Note that the \lstinline{right} combinator is \lstinline{start}ed repeatedly,
once for each leaf node of \lstinline{left}. In general, each
combinator can be managing multiple subtrees of the search.

Multiple \lstinline{andc} combinators may be handling a search
node at the same time. For instance in a heuristic of the form
$\textsf{and}(\textsf{and}(s_1,s_2),s_3)$, two \lstinline{andc} combinators are
active at the same time. The scoping of the associated variables
works in such a way that each \lstinline{andc} has its own instance of
\lstinline{inLeft} for each node.

\subsection{Reusable Combinators}

Now we show how a \emph{monolithic} combinator can be decomposed into more
primitive combinators that can be reused for other purposes.
\pparagraph{Monolithic Combinator} We start from the following
\lstinline{limitsolutions} combinator that prunes the search after
\lstinline{cutoff} solutions have been found. One new concept is the notion of
a global variable associated with a (sub)tree: all descendants of \lstinline{root}
(implicitly) share the same instance of \lstinline{count}. Hence, any update of 
\lstinline{count} by one node is seen by all other nodes in the (sub)tree.

\begin{lstlisting}[language=pseudo,frame=single]
combinator limitsolutions(cutoff,child)
  global int count

  start(root):
    root.count = 0
    child.start(root)

  enter(c):
    if count == cutoff
      parent.exit(abort)
    else
      child.enter(c)

  exit(c,status):
    if status==success 
      c.count++
    parent.exit(c,status)
\end{lstlisting}

\begin{figure}[t]
\begin{center}
\pgfimage[width=0.45\columnwidth]{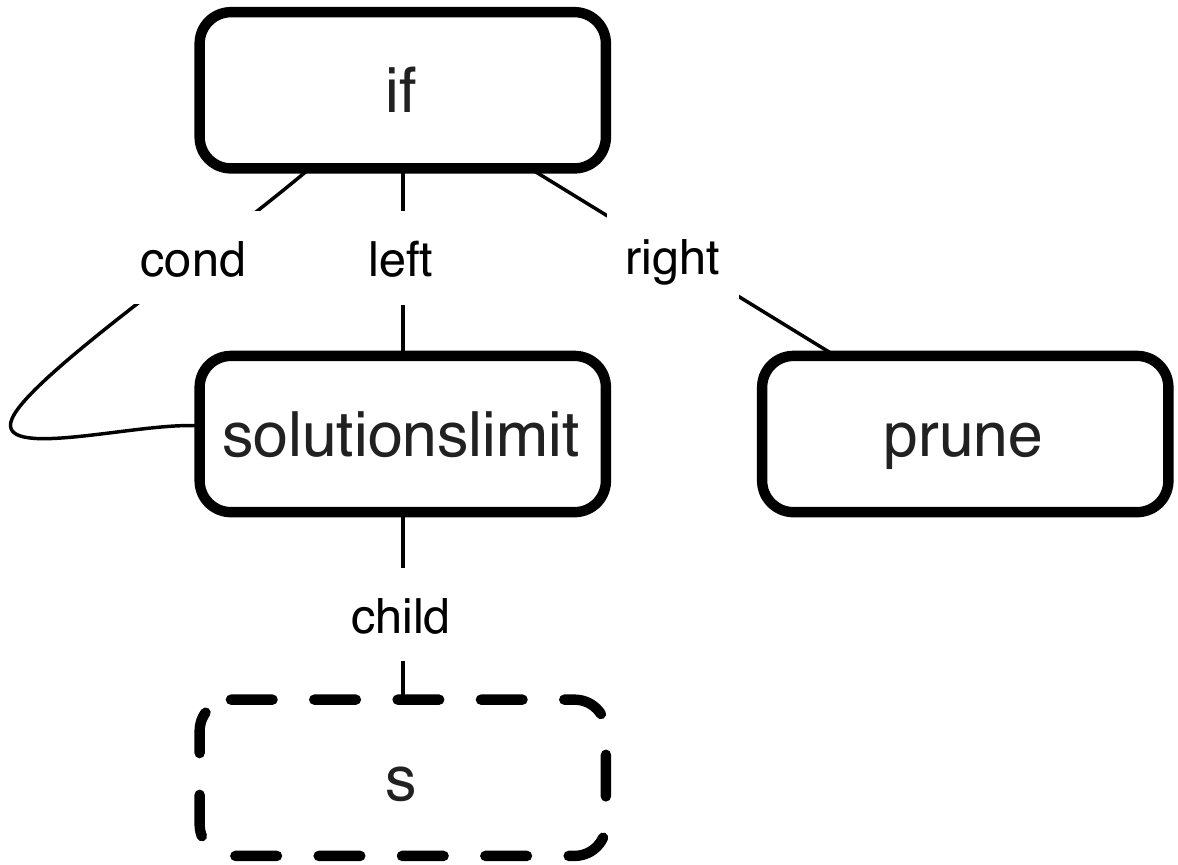}
\end{center}
\caption{The decomposition of the \textsf{limitsolutions} combinator}\label{f:limitsolutions}
\end{figure}

\pparagraph{Decomposition} We can split up the above \lstinline{limitsolutions}
combinator into three different combinators: \lstinline{ifthenelse}, \lstinline{solutionslimit}
and \lstinline{prune}. They form a directed acyclic graph as depicted in Figure~\ref{f:limitsolutions}
or denoted as an expression with sharing below:
\begin{center}
\lstinline{limitsolutions(cutoff,s)} = \lstinline{ifthenelse(s',s',prune)}
\end{center}
where
\begin{center}
\lstinline{s'} = \lstinline{solutionslimit(cutoff,s)}
\end{center}
Here, \lstinline{solutionslimit} monitors the number of solutions produced by \lstinline{s}.
If this number reaches the \lstinline{cutoff}, then \lstinline{ifthenelse} switches to \lstinline{prune},
which discards the remaining nodes in the tree.

We now elaborate on each of these combinators individually.

\pparagraph{Prune} The \textsf{prune} combinator is a minimal base combinator
that immediately exits every node with the \textbf{\lstinline{abort}} status.
The \lstinline{start} message is empty, and the \lstinline{exit} and \lstinline{init}
messages are never called.
\begin{lstlisting}[language=pseudo,frame=single]
combinator prune()
  enter(c):
    parent.exit(c,abort)
\end{lstlisting}

\pparagraph{Solutions Count} The \textsf{solutionslimit} combinator below
illustrates how statistics gathering combinators are implemented. It implements
a sub-protocol of \lstinline{combinator} with an extra message \lstinline{eval}
that queries the current Boolean value:

\begin{minipage}{\textwidth}
\begin{lstlisting}[language=pseudo,xleftmargin=3cm]
protocol condition extends combinator
  eval(currentNode);
\end{lstlisting}  
\end{minipage}

In the case of \lstinline{solutionslimit}, the returned Boolean value is whether
a particular number (\lstinline{cutoff}) of solutions has not yet been reached by
its \lstinline{child}. For this purpose it maintains the number of solutions
found so far in a global variable.

\begin{lstlisting}[language=pseudo,frame=single]
condition solutionslimit(cutoff,child)
  global int count

  start(root):
    root.count = 0
    child.start(root)

  exit(c,status):
    if status==success 
      c.count++
    parent.exit(c,status)

  eval(c): 
    return c.count <= cutoff
\end{lstlisting}

\pparagraph{Ifthenelse} The \lstinline{ifthenelse} combinator is parametrized by one \lstinline{condition} and
two child combinators. It associates with every node whether it is handled by
the \lstinline{left} child (\lstinline{inLeft}); this is the case for the
\lstinline{root} node.  Whenever a node \lstinline{c} is entered that is
\lstinline{inLeft}, the condition is checked.  If the condition fails, \lstinline{c}
becomes the root of a subtree that is further handled by \lstinline{right}.

\begin{lstlisting}[language=pseudo,frame=single]
combinator ifthenelse(cond,left,right)
  local bool inLeft
  
  start(root):
    root.inLeft=true
    left.start(root)
  
  enter(c):
    if not c.inLeft 
      right.enter(c)
    else if cond.eval() 
      left.enter(c)
    else 
      c.inLeft=false
      right.start(c)
      right.enter(c)

  init(p,c):
    c.inLeft=p.inLeft
    if c.inLeft 
      left.init(p,c) 
    else 
      right.init(p,c)
\end{lstlisting}

\subsection{Restarting Combinators}

Restarting the search is common to several combinators; the mechanic is illustrated
below in the \textsf{portfolio} combinator.

\pparagraph{Portfolio}
Like the \textsf{ifthenelse} and \textsf{and} combinators, the \textsf{portfolio}
combinator switches between child combinators. Only the logic for switching
is more complex. In order to simplify presentation, we again restrict the code to the binary case; the $n$-ary variant is a straightforward generalization.

Firstly, \textsf{portfolio} keeps track of a global ``reference'' count \texttt{ref} of
unprocessed nodes to be handled by the \texttt{s1} child. This count is
incremented whenever a new child node is initialized, and decremented whenever
a node is entered for actual processing.

When the last node of \texttt{s1} exits (witnessed by the reference count being $0$) and the search was not exhaustive, the
search starts over from the root, but now with the \texttt{s2} child.  In
order to decide about exhaustiveness, the \textsf{portfolio} combinator
registers whether any \lstinline{exit} with status \lstinline{abort} occurred.  At
the same time it converts an \lstinline{abort} inside \texttt{s1} into a
\texttt{failure}, because the \texttt{s2} combinator may still perform an
exhaustive search and avoid overall non-exhaustiveness.
In order to restart from the \texttt{root}, a \texttt{copy} of the
\texttt{root} node is made at the \lstinline{start}. 

Upon a successful \lstinline{exit}, the leaf node becomes the root
of a new subtree that is further handled by the \texttt{s2} combinator.
\begin{lstlisting}[language=pseudo,frame=single]
combinator portfolio(s1,s2)
  global node copy
  global bool inLeft
  global bool exhaustive
  global int ref

  start(root):
    copy=root.copy()
    root.inLeft=true
    root.exhaustive=true
    root.ref=1
    s1.start(root)
  
  enter(c):
    if c.inLeft 
      ref--
      s1.enter(c)
    else 
      s2.enter(c)
  
  exit(c,status):
    if not c.inLeft 
      parent.exit(c,status)
    else
      if status==abort 
        status=failure
        c.exhaustive=false
      if c.ref==0 
        if c.exhaustive 
          parent.exit(c,status)
        else 
          copy.inLeft=false
          s2.start(copy)
          self.enter(copy)
      else 
        parent.exit(c,status)

  init(p,c):
    ref++;
    if c.inLeft 
      s1.init(p,c) 
    else 
      s2.init(p,c)
\end{lstlisting}

\section{Modular Combinator Implementation}
\label{sec:modular_combinator_implementation}

The message-based combinator approach lends itself well to different
implementation strategies. In the following we briefly discuss two
diametrically opposed approaches we have explored:  \textit{dynamic
composition} (interpretation) and \textit{static composition} (compilation).
Using these different approaches, combinators can be adapted to the implementation choices of
existing solvers. Sect.~\ref{sec:experiments} shows that both implementation approaches have
competitive performance.

\subsection{Dynamic Composition}

To support dynamic composition, we have implemented our combinators as \CPP{}
classes whose objects can be allocated and composed into a search specification at runtime. The
protocol events correspond to virtual method calls between these objects. For the delegation mechanism from
one object to another, we explicitly encode a form of dynamic inheritance
called \textit{open recursion} or \textit{mixin inheritance}~\cite{cook}. 
In contrast to the OOP inheritance built into \CPP{} and Java, this mixin
inheritance provides two essential abilities: 1) to determine the inheritance graph \textit{at runtime}
and 2) to use multiple copies of the same combinator class at different points in the inheritance graph.
In contrast, \CPP{}'s built-in static inheritance provides neither.

The \CPP{} library currently builds on top of the Gecode constraint solver~\cite{gecode}.
However, the solver is accessed through a layer of abstraction that is
easily adapted to other solvers (e.g., we have a prototype interface to the 
Gurobi MIP solver). The complete library weighs in at around 2500
lines of code, which is even less than Gecode's native search and branching
components.

\subsection{Static Composition}

In a second approach, also on top of Gecode, we statically compile a search specification to a tight
\CPP{} loop. Again, every combinator is a separate module independent of other combinator modules. A combinator module now does not directly implement the
combinator's behavior. Instead it implements a code generator (in Haskell),
which in turn produces the \CPP{} code with the expected behavior.

Hence, our search language compiler parses a search specification, and composes
(in mixin-style) the corresponding code generators. Then it
runs the composite code generator according to the message protocol.
The code generators produce appropriate \CPP{} code fragments for the different
messages, which are combined according to the protocol into the monolithic
\CPP{} loop.
This \CPP{} code is further post-processed by the
\CPP{} compiler to yield a highly optimized executable. 

As for dynamic composition, the mixin approach is crucial, 
allowing us to add more combinators without touching the
existing ones. At the same time we obtain with the press of a button
several 1000 lines of custom low-level code for the composition of just a few
combinators. In contrast, the development cost of hand crafted code is
prohibitive. 

\paragraph{A compromise} between the above two approaches, itself static, is to employ the
built-in mixin mechanism (also called \textit{traits}) available in
object-oriented languages such as Scala~\cite{scala:typesystem} to compose
combinators. A dynamic alternative is to generate the combinator
implementations using dynamic compilation techniques, for instance using the
LLVM (Low Level Virtual Machine) framework. These options remain to be
explored.

\subsection{MiniZinc with Combinators}
\label{sec:minizinccomb}

\lstdefinelanguage{zinc} 
  {morekeywords={annotation,var,int,ann,solve,satisfy}
  ,classoffset=1
  ,sensitive=false
  ,morecomment=[l]{//}
  ,morecomment=[s]{/*}{*/}
  ,morestring=[b]"
  ,literate=
  }

As a proof of concept and platform for experiments, we have integrated search combinators into a complete MiniZinc toolchain, comprising a pre-compiler and a FlatZinc interpreter.

The pre-compiler is necessary to support arbitrary expressions in annotations, such as the condition expressions for an \textsf{ifthenelse}. The expressions are translated into standard MiniZinc annotations that are understood by the FlatZinc interpreter. User-defined variables have type-inst \texttt{svar int} and can be introduced using the standard MiniZinc \texttt{let} construct. The \texttt{annotation} construct of MiniZinc has been extended to support simple function definitions. The following example shows a MiniZinc version of the restart-based branch-and-bound heuristic from Sect.~\ref{sec:stateaccess}:

\begin{minipage}{\textwidth}
\begin{lstlisting}[language=zinc]
annotation limit(var bool: c, ann: s) =
    ifthenelse(c,s,prune);

annotation once(ann: s) = limit(solutions < 1, s);

annotation rbab(var int: obj, ann: s) =
  let { svar int: best = MAXINT } in
  restart(true, and([
            post(obj < best),
            once(s),
            assign(best,obj)]));

solve ::rbab(x,int_search(y,input_order,assign_lb)) satisfy;
\end{lstlisting}
\end{minipage}

The pre-compiler translates this code as follows:

\begin{minipage}{\columnwidth}
\begin{lstlisting}[language=zinc]
solve :: sh_let(sh_letvar("best"), sh_int(MAXINT),
  sh_restart(sh_cond_true, sh_and([
    sh_post_succeed(sh_cond_lt(sh_intvar(objective),
                               sh_letvar("best"))),
    sh_let(sh_letvar("solutioncount"), 0,
      sh_ifthenelse(sh_cond_lt(sh_letvar("solutioncount"),
                               sh_int(1)),
        sh_solutioncount(sh_letvar("solutioncount"),
          sh_int_search(x, sh_var_input_order,
                           sh_val_assign_lb)), 
        sh_prune)),
    sh_assign(sh_letvar("best"), sh_intvar(objective))])))
    satisfy;
\end{lstlisting}
\end{minipage}

All literals are quoted (e.g.\ \lstinline{sh_int(1)}), user-defined search variables are turned into quoted  strings (\lstinline{lv("best")}), expressions like \lstinline{obj < best} are translated into annotation terms (\lstinline{sh_cond_lt} \dots), and statistics are made explicit, introducing search variables and special combinators (\lstinline{sh_solutioncount}). The result of the pre-compilation is valid, well-typed MiniZinc, which is then passed through the standard \texttt{mzn2fzn} translator to produce FlatZinc ready for solving. We intend to incorporate the translations done by the pre-compiler into the standard \texttt{mzn2fzn} in the future.

We extended the Gecode FlatZinc interpreter to parse the search combinator annotation and construct the corresponding heuristic using the Dynamic Composition approach described above. The three tools, pre-compiler, \texttt{mzn2fzn}, and the modified FlatZinc interpreter thus form a complete toolchain for solving MiniZinc models using search combinators. The source code including examples can be downloaded from\\
\url{http://www.gecode.org/flatzinc.html}.

\section{Experiments}
\label{sec:experiments}

This section evaluates the performance of our two implementations. It
establishes that a search heuristic specified using combinators is competitive
with a custom implementation of the same heuristic, exploring exactly the same
tree.

Sect.~\ref{s:design:protocol} introduced a message protocol that defines the
communication between the different combinators \emph{for one node of the
search tree}. Any overhead of a combinator-based implementation must therefore
come from the processing of each node using this protocol. All combinators
discussed earlier process each message of the protocol in constant time (except
for the \textsf{base\_search} combinators, of course). Hence, we expect at
most a constant overhead per node compared to a native
implementation of the heuristic.

In the following, two sets of experiments confirm this expectation. The first set consists of artificial benchmarks designed to expose the overhead per node. The second set consists of realistic combinatorial problems with complex search strategies.

The experiments were run on a 2.26~GHz Intel Core~2 Duo running Mac~OS~X. The results are the averages of 10 runs, with a coefficient of deviation less than 1.5\%.

\pparagraph{Stress Test}
The first set of experiments measures the overhead of calling a single
combinator during search. We ran a complete search of a tree generated by 7
variables with domain $\{0,\dots,6\}$ and \emph{no} constraints (1\,647\,085
nodes). To measure the overhead, we constructed a basic search heuristic $s$
and a stack of $n$ combinators:

$\portfolio{[\portfolio{[\dots\portfolio{[s,\failure{}]}\dots,\failure{}]},\failure{}]}$

\noindent
where
$n$ ranges from $0$ to $20$ (realistic combinator stacks, such as
those from the examples in this article, are usually not deeper than 10).  The
numbers in the following table report the runtime with respect to using the
plain heuristic $s$, for both the static and the dynamic approach:
{\small
\begin{center}
\begin{tabular}{l|@{\hspace{5mm}}r@{\hspace{5mm}}r@{\hspace{5mm}}r@{\hspace{5mm}}r@{\hspace{5mm}}r}
  $n$ & 1 & 2 & 5 & 10 & 20 \\\hline
  \emph{static \%} & 106.6 & 107.7 & 112.0 & 148.3 & 157.5\\
  \emph{dynamic \%} & 107.3 & 117.6 & 145.2 & 192.6 & 260.9
\end{tabular}  
\end{center}
}

\noindent
A single combinator generates an overhead of around 7\%, and 10 combinators add 50\% for the static and 90\% for the dynamic approach. In absolute runtime, however, this translates to an overhead of around 17 ms (70 ms) per million nodes and combinator for the static (dynamic) approach. Note that this is a worst-case experiment, since there is no constraint propagation and almost all the time is spent in the combinators.

\pparagraph{Benchmarks}
The second set of experiments shows that in practice, this overhead is dwarfed
by the cost of constraint propagation and backtracking. Note that the
experiments are not supposed to demonstrate the best possible search heuristics
for the given problems, but that a search heuristic implemented using
combinators is just as efficient as a native implementation.

Fig.~\ref{tab:golomb} compares Gecode's optimization search engines with
branch-and-bound implemented using combinators. On the well-known Golomb Rulers
problem, both dynamic combinators and native Gecode are slightly slower than
static combinators. Native Gecode uses dynamically combined search heuristics,
but is much less expressive. That is why the
static approach with its specialization yields better results.

On the radiotherapy problem (see Sect.~\ref{sec:stateaccess}), the dynamic combinators show an overhead of 6--11\%. For native Gecode, \textsf{exh\_once} must be implemented as a nested search, which performs similarly to the dynamic combinators. However, in instances 5 and 6, the compiled combinators lose their advantage over native Gecode. 
This is due to the processing of \textsf{exh\_once}: As soon as it is finished, the combinator approach processes all nodes of the \textsf{exh\_once} tree that are still in the search queue, which are now pruned by \textsf{exh\_once}. The native Gecode implementation simply discards the tree. We will investigate how to incorporate this optimization into the combinator approach.

The job shop scheduling examples, using the combination of branch-and-bound and discrepancy limit discussed in Sect.~\ref{sec:stateaccess}, show similar behavior. In ABZ1-5 and mt10, the interpreted combinators show much less overhead than in the short-running instances. This is due to more expensive propagation and backtracking in these instances, reducing the relative overhead of executing the combinators.

In summary, the experiments show that the expressiveness and flexibility of a rich combinator-based search language can be achieved without any runtime overhead in the case of the static approach, and little overhead for the dynamic version.

\begin{figure}[t]
\begin{center}
\begin{tabular}{l|@{\hspace{5mm}}r@{\hspace{5mm}}r@{\hspace{5mm}}r}
  & \emph{Compiled} & \emph{Interpreted} & \emph{Gecode} \\\cline{1-4}

\emph{Golomb 10} & 0.61 s & 101.8\% & 102.5\%\\
\emph{Golomb 11} & 12.72 s & 102.9\% & 101.8\%\\
\emph{Golomb 12} & 125.40 s & 100.6\% & 101.9\%\\\cline{1-4}
\emph{Radiotherapy 1} & 71.13 s & 105.9\% & 107.3\% \\
\emph{Radiotherapy 2} & 6.22 s & 110.9\% & 110.1\% \\
\emph{Radiotherapy 3} & 11.78 s & 108.3\% & 108.1\% \\
\emph{Radiotherapy 4} & 16.44 s & 107.5\% & 106.9\% \\
\emph{Radiotherapy 5} & 69.89 s & 108.1\% & 98.7\% \\
\emph{Radiotherapy 6} & 106.04 s & 109.2\% & 99.1\%\\\cline{1-4}
\emph{Job Shop G2} & 7.25 s      & 146.3\% & 101.2\% \\
\emph{Job-Shop G4} & 6.96 s      & 164.0\% & 107.75\% \\
\emph{Job-Shop H1} & 38.05 s     & 153.1\% & 103.81\% \\
\emph{Job Shop H3} & 52.02 s     & 162.5\% & 102.8\% \\
\emph{Job Shop H5} & 20.88 s     & 153.2\% & 107.0\% \\
\emph{Job Shop ABZ1-5} & 2319.00 s  & 103.7\% & 100.1\%\\
\emph{Job Shop mt10} & 2181.00 s    & 104.5\% & 99.9\%
\end{tabular}
\end{center}
  \caption{Experimental results}
  \label{tab:golomb}
\end{figure}

\section{Related Work}
\label{sec:related_work}

This section explores and discusses previous work that is closely related to search combinators as presented in this article.

\subsection{MCP}

This work directly extends our earlier work on 
\textbf{Monadic Constraint Programming} (MCP) \cite{mcp}. 
MCP introduces stackable search transformers, which are a simple form
of search combinators, but only provide a much more limited and low level
form of search control.  In trying to overcome its limitations we arrived
at search combinators.

\subsection{Constraint Logic Programming}

Constraint logic programming languages such as \textbf{ECLiPSe}~\cite{eclipse}
and \textbf{SICStus} Prolog~\cite{sicstus} provide programmable search via the
built-in search of the paradigm, allowing the user to define \emph{goals} in
terms of conjunctive or disjunctive sub-goals. 

Prolog's limitation is that it does not permit cross-cutting between goals.
For instance, disjunctions inside goals are too well encapsulated to observe
them or interfere with them from outside that goal. Hence, combinators that
inject additional behavior in disjunctions, i.e. to observe and/or prune the
number of branches, cannot be expressed in a modular way. In contrast,
cross-cutting is a crucial feature of our combinator approach, where a
combinator higher up in the stack can interfere with a sub-combinator, while
remaining fully compositional. In summary, apart from conjunction and
disjunction, Prolog's goal-based heuristics cannot be combined arbitrarily. 

ECLiPSe copes with this limitation by combining a limited number of search
heuristics into a monolithic \texttt{search/6} predicate.  With various
parameters the user controls which of the heuristics is enabled (e.g.,
depth-bounded, node-bounded or limited discrepancy search). A fixed number of
compositions are supported, such as changing strategy
when the depth bound finishes. The labeling itself is user programmable.
If a user is not happy with the set of supported heuristics in \texttt{search/6},
he has to program his own from scratch.

\textbf{IBM ILOG CP Optimizer}~\cite{cpoptimizer} supports Prolog-style
goals in \CPP~\cite{Puget:1994:0:A-CPP-Implementation}, 
and like Prolog goals, these do not support cross-cutting.

\subsection{The Comet Language}

The \textbf{Comet}~\cite{comet} system features fully programmable
search~\cite{cometsearch}, built upon the basic concept of
\emph{continuations}, which make it easy to capture the state of the solver and
write non-deterministic code.

The Comet library provides abstractions like the non-deterministic primitives
\texttt{try} and \texttt{tryall} that split the search specification in two
(orthogonal) parts: 1) the specification of the search tree which corresponds to our to our
\textsf{base\_search} heuristics, and 2) the exploration of that search tree by
means of a \emph{search controller}.
In terms of our approach, the search controller determines both the queueing
strategy and the behavior of the search heuristic (minus the base search)
within a single entity. In other words, it defines what to do when starting or
ending a search, failing, or adding a new choice. 

Complex heuristics can be constructed as
custom controllers, either by inheriting from existing controllers or
implementing them from scratch.

Albeit at a different level of abstraction (e.g., compare the Comet definition
of depth-bounded search in Figure~\ref{f:comet:dbs} to the combinator definition
$\textsf{dbs}(n,s) \equiv \textsf{limit}(\textsf{depth} \leq n, s)$),
search controllers are quite similar to combinators as presented in this
article. However, there is one essential difference. Our combinators are meant to be
compositional, whereas search controllers are not. This difference in spirit is
clearly reflected in 1) the design of the interface and its associated
protocol, and 2) the instances:

\begin{enumerate}
\item
The design of search controllers is simpler than that of search combinators
because it does not take compositionality into account. While many of the
messages in the two approaches are similar in spirit, the search combinator
approach also stipulates the flow of messages within a search combinator
composition. Notably, while most of the messages propagate top-down through a
combinator stack, it is vital to compositionality that the \lstinline{exit}
message proceeds in a bottom-up manner. For instance, this bottom-up flow
enables the inner \textsf{and} combinator in the composition
$\textsf{and}(\textsf{and}(s_1,s_2),s_3)$ to intercept leaf nodes of $s_1$ and
start $s_2$ before its parent starts $s_3$. The other way around would clearly
exhibit an undesirable semantics.

In Comet, this compositional protocol is entirely absent. All messages are
directed at the single search controller.

\item
In terms of instantiation, because of their compositional nature, we promote
many ``small'' combinator instances that each capture a single primitive
feature.  This approach provides us with a high-level modeling language for
search, as the primitive combinators are conveniently assembled into many
different search heuristics. In contrast, all Comet search controller instances
we are aware of\footnote{i.e., those published in papers and shipped with the
Comet library.} are essentially monolithic implementations of a particular
search heuristic; none of them takes other search controllers as arguments.
Through a common abstract base class the instances share some basic
infrastructure, but to implement a new search controller one basically starts
from scratch.
\end{enumerate}

The fact that search controllers have not been designed with compositionality
in mind obviously does not mean that compositionality cannot be achieved in
Comet.  On the contrary, we believe that it is most easily achieved by
integrating search controllers with the compositional design of our search
combinators.  In fact, because of Comet's powerful primitives for
non-determinism, this would lead to a particularly elegant implementation.

\lstdefinelanguage{Comet} 
  {morekeywords={class,implements,int,new,void,do,while,if,else,super,call,bool,extends,Continuation,stack}
  ,classoffset=1
  ,morekeywords={start,fail,exit,addChoice,startTry}
  ,keywordstyle=\underline
  ,sensitive=false
  ,morecomment=[l]{//}
  ,morecomment=[s]{/*}{*/}
  ,morestring=[b]"
  ,literate=
  }
\begin{figure}
\begin{lstlisting}[language=Comet]
class DBS extends AbstractSearchController {
    stack{Continuation} s;
    int limit;
    DBS(SearchSolver so, int n) : AbstractSearchController(so) { 
        s = new stack{Continuation}();
        limit = n;
    } 
    void startTry() {
        if (s.getSize() > limit) fail();
    }
    void addChoice(Continuation f) {
        s.push(f);
    } 
    void fail() {
        if (s.empty())
            exit();
        else 
         call(s.pop());
    }
}
\end{lstlisting}
\caption{Definition of depth-bounded search in Comet.}\label{f:comet:dbs}
\end{figure}

\subsection{Other Systems}

The \textbf{Salsa}~\cite{salsa} language is an imperative domain-specific
language for implementing search algorithms on top of constraint solvers. Its
center of focus is a node in the search process. Programmers can write custom
\emph{Choice} strategies for generating next nodes from the current one;
Salsa provides a regular-expression-like language for combining these
Choices into more complex ones. In addition, Salsa can run custom
procedures at the \textit{exits} of each node, right after
visiting it. We believe that Salsa's Choice construct is orthogonal 
to our approach and could be incorporated. Custom exit procedures 
show similarity to combinators, but no support is provided for 
arbitrary composition.

\textbf{Oz}~\cite{oz} 
was the first language to truly separate the 
definition of the constraint model 
from the exploration strategy~\cite{ozsearch}. 
Computation spaces capture the solver state and the possible
choices. Strategies such as DFS, BFS, LDS, Branch and Bound and Best 
First Search are implemented by a combination of \emph{copying} and \emph{recomputation} of computation spaces.
The strategies are monolithic, there is no notion of
search combinators.

\textbf{Zinc/MiniZinc}~\cite{zinc,minizinc} lets the user specify search in
its \emph{annotation language}. There is a proposal for a more expressive
search language for MiniZinc~\cite{samulowitz:2010:0:towards}, but it is
limited to basic variable ordering and domain splitting strategies. 
For Zinc, a language extension is available for 
implementing variable selection and domain 
splitting~\cite{adding_search_to_zinc} 
but again it does not address more than basic search.

\subsection{Autonomous Search}

Autonomous search (AS)~\cite{as} addresses the challenge of providing complex
application-tailored search heuristics in a different way. Rather than leaving the
specification and tuning of search heuristics to the programmer, AS promotes
systems that autonomously self-tune their performance while solving problems.
Hence, while search combinators make writing search heuristics easier, AS takes
it out of the hands of the programmer altogether. Well-known instances of this approach are Impact Based Search~\cite{impact} or the \emph{weighted degree} heuristic~\cite{wdeg}.

AS has advantages for 1) smaller problems where it produces a decent heuristic without programmer
investment, and for 2) novice users who don't know how to obtain a decent
heuristic. However, loss of programmer control is a liability for hard problems
where AS can be ineffective and often only expert knowledge makes the
difference.

\section{Conclusion}
\label{sec:conclusion}

We have shown how combinators provide a powerful high-level
language for modeling complex search heuristics. To make this approach useful in practice, we devised an architecture in which the modularity of the language is matched by the modularity of the implementation. This
relieves system developers from a high implementation cost and yet, as our experiments show, imposes no runtime penalty.

For future work, parallel search on multi-core hardware fits perfectly
in our combinator framework. We have already performed a number of preliminary
experiments and will further explore the benefits of search
combinators in a parallel setting. We will also explore potential optimizations (such as the short-circuit of \textsf{exh\_once} from Sect.~\ref{sec:experiments}) and different compilation strategies (e.g., combining the static and dynamic approaches from Sect.~\ref{sec:modular_combinator_implementation}). 

In addition we consider to apply search combinators in other problem domains like Mixed Integer Programming (MIP) and 
$A^*$ where search strategies have a major impact on performance and no dominant default search exists.
Finally, combinators need not necessarily be heuristics that control
the search. They may also monitor search, e.g., by gathering
statistics or visualizing the search tree. 

\begin{acknowledgements}
NICTA is funded by the Australian Government as represented by the 
Department of Broadband, Communications and the Digital Economy and the 
Australian Research Council. 
This work was partially supported by Asian Office of Aerospace Research
and Development grant 10-4123.  
\end{acknowledgements}

\bibliographystyle{spmpsci} \bibliography{biblio}

\end{document}